# Asking the Right Questions: Benchmarking Large Language Models in the Development of Clinical Consultation Templates

**Liam G. McCoy†**
*Division of Neurology, Faculty of Medicine and Dentistry, University of Alberta, Edmonton, AB, Canada; Department of Medicine, Beth Israel Deaconess Medical Center, Boston, MA, USA;*
*Email: lmccoy@ualberta.ca*

**Fateme Nateghi Haredasht† and Kanav Chopra**
*Stanford Center for Biomedical Informatics Research*
*Stanford, CA, USA*
*Email: fnateghi@stanford.edu; kanav@stanford.edu*

**David Wu**
*Department of Dermatology, Mass General Brigham, Harvard Medical School, Boston, MA, USA*
*Email: dwu@mgh.harvard.edu*

**David JH Wu and Abass Conteh**
*Department of Radiation Oncology, Stanford Cancer Center, Palo Alto, CA, USA*
*Email: davidjwu@stanford.edu; aconteh@stanford.edu*

**Sarita Khemani and Saloni Kumar Maharaj**
*Division of Hospital Medicine, Stanford University School of Medicine, Stanford, CA, USA*
*Email: skhemani@stanford.edu; salonikumar@stanford.edu*

**Vishnu Ravi**
*Stanford Mussallem Center for Biodesign, Stanford University; Stanford University School of Medicine, Stanford, CA, USA*
*Email: vishnur@stanford.edu*

**Arth Pahwa**
*Division of Neurology, Faculty of Medicine and Dentistry, University of Alberta, Edmonton, AB, Canada*
*Email: apahwa@ualberta.ca*

**Yingjie Weng**
*Quantitative Sciences Unit, Stanford University School of Medicine, Stanford, CA, USA*
*Email: yweng7@stanford.edu*

**Kelvin Zhenghao Li**
*Department of Ophthalmology, Stanford Byers Eye Institute; Department of Ophthalmology, Tan Tock Seng Hospital; Centre of AI in Medicine, Lee Kong Chian School of Medicine, Nanyang Technological University Palo Alto, CA, USA; Singapore, Singapore*
*Email: kelvinli@stanford.edu*

**Leah Rosengaus and Lena Giang**
*Stanford Health Care, Stanford Medicine, Palo Alto, CA, USA*
*Email: LRosengaus@stanfordhealthcare.org; LGiang@stanfordhealthcare.org*

**Olivia Jee**
*Division of Primary Care and Population Health, Stanford University School of Medicine, Palo Alto, CA, USA*
*Email: ojee@stanford.edu*

**Daniel Shirvani**
*Faculty of Medicine, University of British Columbia, Vancouver, BC, Canada*
*Email: dshirv@student.ubc.ca*

**Ethan Goh***
*Stanford Center for Biomedical Informatics Research Clinical Excellence Research Center, Stanford School of Medicine,*
*Stanford, CA, USA*
*Email: ethangoh@stanford.edu*

**Jonathan H. Chen***
*Stanford Center for Biomedical Informatics Research Division of Hospital Medicine, Stanford School of Medicine; Clinical Excellence Research Center, Stanford School of Medicine; Department of Medicine, Stanford University,*
*Stanford, CA, USA*
*Email: jonc101@stanford.edu*

**†Authors contributed equally. *Authors jointly supervised this work. Corresponding: Ethan Goh**

**Abstract:**

This study evaluates the capacity of large language models (LLMs) to generate structured clinical consultation templates for electronic consultation. Using 145 expert-crafted templates developed and routinely used by Stanford's eConsult team, we assess frontier models—including o3, GPT-4o, Kimi K2, Claude 4 Sonnet, Llama 3 70B, and Gemini 2.5 Pro—for their ability to produce clinically coherent, concise, and prioritized clinical question schemas. Through a multi-agent pipeline combining prompt optimization, semantic autograding, and prioritization analysis, we show that while models like o3 achieve high comprehensiveness (up to 92.2%), they consistently generate excessively long templates and fail to correctly prioritize the most clinically important questions under length constraints. Performance varies across specialties, with significant degradation in narrative-driven fields such as psychiatry and pain medicine. Our findings demonstrate that LLMs can enhance structured clinical information exchange between physicians, while highlighting the need for more robust evaluation methods that capture a model's ability to prioritize clinically salient information within the time constraints of real-world physician communication. Limitations include reliance on Stanford-specific templates and concordance-based grading, which may not capture all clinically reasonable outputs.

## 1. Introduction

Medicine is fundamentally a knowledge processing and collaborative discipline[1]. High-quality consultation between generalist clinicians and specialists ensures accurate, timely, and efficient patient care. Communication breakdowns contribute frequently to medical errors[2], leading to death and preventable harm[3–5]. Improving the structure and quality of specialist–generalist consultation templates offers a direct opportunity to reduce diagnostic delays, treatment errors, and unnecessary testing[6]. Effective specialist consultations hinge on a subtle cognitive balance—comprehensively capturing clinical details, while anticipating and selectively prioritizing the specialist's informational needs[7,8]. Traditional synchronous consultations, such as those conducted over the phone, enable specialists to ask targeted follow-up questions, dynamically acquiring necessary details to inform their clinical decision.

Asynchronous e-consultation systems offer a structured workflow to streamline primary care physician (PCP)-specialist interactions but place greater reliance on the initial information provided by the consulting physician[9–12]. Stanford's SAGE eConsult product (Specialist AI Guiding Experts) utilizes specialist-generated consultation templates (*Supplementary Figure S1*) developed by Stanford's eConsult team to guide primary care physicians in submitting structured, high-yield clinical consults[13]. Each template outlines both required and optional elements—such as key history, medications, labs, and comorbidities—that enable specialists to provide more actionable recommendations. However, maintaining a library of expert-crafted templates is resource-intensive, requiring continuous updates to reflect evolving guidelines, specialist expectations, and health system workflows. Static templates also fail to capture the breadth of clinical variability and edge cases[14,15]. Given the wide variability of possible conditions and relevant questions, novel AI template generation is important to fully scale this workflow.

We evaluate the ability of contemporary large language models (LLMs) to generate clinically useful consultation templates that can contextually adapt to individual consults, expanding coverage beyond common scenarios while preserving clinical rigor and relevance. Beyond its practical value, template generation represents a unique cognitive challenge rarely examined in medical natural language processing (NLP): models must anticipate what another clinician needs to know, not simply recall facts or summarize text. Evaluating this capability required developing new approaches that measure performance under the competing demands of comprehensiveness and conciseness, challenges absent from traditional benchmarks. Through this work, we aim to highlight both the practical clinical potential and the technical challenges in deploying LLMs for nuanced clinical communication tasks.

## 2. Methods

### 2.1. *Data*

The Stanford eConsult Templates dataset consists of 173 expert-generated templates for inbound specialist consultations generated from 2019 to 2025. These templates are based on frequently received consults and consist of mandatory and recommended information for the consulting physician to include. Templates were screened for inclusion based on clinical relevance and specificity. Exclusion criteria included: (i) templates that consisted primarily of redirection to other services, (ii) templates that consisted exclusively of links to educational resources or instructions for photo submission (most notably in dermatology), and (iii) "other" templates that did not specify a condition. After applying these criteria, the final dataset consisted of 145 templates across 20 specialties (allergy, cardiology, hematology, chemical dependency, endocrinology, otolaryngology, gastroenterology, gynecology, infectious disease, interventional pulmonology, LGBTQ+ medicine, nephrology, neurology, orthopedics, pain medicine, psychiatry, pulmonology, rheumatology, sleep medicine, and urology). These templates are highly variable in content, containing 1601 unique fields, with 99.3% of these fields present in <10% of templates, and an average Jaccard similarity of 23.7% within specialties. An example template can be found in *Supplementary Figure S1*.

### 2.2. *Response Autograder*

LLM-generated template items rarely match the expert templates verbatim. Therefore, we deployed a semantic autograder that judges clinical equivalence at the item level (*Figure 1)*. The grader is an OpenAI o3[16] instance prompted to decide, for every pair of strings, whether the LLM-generated item satisfies the information request of the expert template item. For example, a generation of "TSH" is accepted as equivalent to the expert template item of "thyroid function tests". Concordance analysis was performed on a per-item basis, assessing whether the item in the original template has an equivalent in the generated template (Figure 3A). Conciseness was evaluated by comparing the total number of novel generated template items to the total number of items in the original template.

The autograder was first calibrated by assessment of a random sample of 40 templates graded by either GPT-4.1[17] or o3. Three blinded US board-certified internal medicine physicians (SK, VR, SM) reviewed the accuracy of the autograder's matches, rating them (i) a full match, if the generated item was clinically equivalent to the original item or fully contained its contents (ii) a partial match, if the generated item included the original item but with a clinically significant difference, and (iii) a disagreement, if the autograder erred in matching the two items. To assess whether the autograder was excessively conservative in its matching, items from the original templates that were rated as 'unmatched' by the model were manually reviewed against all generated items. Final autograder validation was performed, with an iteratively updated prompt, on a set of 20 additional templates evaluated by the same criteria.

Krippendorff's α[18,19] (multi-rater, nominal, tolerant of missingness) and Gwet's AC1[20] (chance-corrected agreement, less sensitive than Cohen's κ to prevalence/marginal imbalance[21]) were used to compare the three clinicians and the autograder. For Krippendorff's α, values ≥0.80 generally indicate strong agreement, whereas ~0.67–0.80 support tentative conclusions[19,22], whereas Gwet's AC1 does not have standard interpretive bands.

## 2.3. *Prompt Optimization*

In our study, Declarative Self-Improving Python (DSPy)[23] was used to generate model-specific prompts for each candidate LLM, refining examples, formatting, and instructions to maximize both clinical quality and structural adherence. DSPy is an open-source framework that treats prompt engineering as a learnable, declarative pipeline. Rather than manually crafting prompts, the user specifies an objective—such as "maximize fidelity to the SAGE template schema under a length constraint"—and DSPy automatically explores variations in prompt structure to optimize performance against this target. Each model was trained on 20 specialty-specific templates, with its outputs evaluated by an o3-based autograder that penalized both omissions of key elements and excessive verbosity. DSPy used this feedback in an iterative loop to adapt the prompt over successive rounds, allowing model performance itself to guide the evolution of the prompt until improvements plateaued. Full prompt details, including examples, may be found at https://github.com/liamgmccoy/stanfordtemplatepublic.

## 2.4. *Template Generation*

Following DSPy-based few-shot prompt optimization using 20 templates, the finalized prompt was applied to the remaining 125 conditions. For each, the model was provided only the specialty and condition and tasked with generating a complete consultation template (*Supplementary Figure S1*)

## 2.5. *Template Prioritization Analysis:*

In order to assess for conciseness as well as comprehensiveness, a secondary prioritizer agent (using the same underlying model, e.g., an o3 prioritizer for o3-generated templates, or a K2 prioritizer for K2 templates) (*Figure 1*) was used to rank the generated template items in order of clinical importance. From this ranked list, truncated sets of the original generated template were produced based on a "top-*n*" format where *n* equals the number of elements in the original template (that is, for an original template with 6 items, top-*n* would take the top 6 elements from the original template, top-*n*+3 would take the top 9 elements, etc). This enables a degree of normalization for the length and complexity of the original templates. These truncated templates were placed through the standard autograder evaluation process, with their relative performance assessed using both standard recall values (recall@k[24], as a percentage of the original template items present within the truncated subset) and normalized discounted cumulative gain at k (NDCG@k)[25], calculated by assigning a value of 2 to items labelled "required" in the initial templates, and 1 to those labelled "optional".

### 2.6. *Model Benchmarking:*

The performance of multiple frontier models on the template generation task was assessed, including OpenAI's GPT-4o[26] and o3[16] models, Google's Gemini 2.5 Pro[27], Anthropic's Claude 4 Sonnet[28], and leading open-source models Kimi K2[29] and Llama 3 70B[30]. Models were compared for both overall template generation performance as well as their performance on the prioritization task

### 2.7. *Novel Template Generation:*

In order to assess the generalization of template generation, novel cases were generated in the specialties of infectious disease, neurology, ophthalmology, and radiation oncology based on five novel topics per specialty suggested by a board-certified (KL) or senior resident physician (DW, AP, AC) co-author in each specialty. All templates were generated by o3 with the DSPy-optimized final prompt. These cases were reviewed item by item for the clinical relevance and importance of each suggested item, with the additional ability for the authors to add additional important items.

## 3. Results

### 3.1. *Novel Template Generation*

Novel templates were generated for 20 expert-selected topics across the specialties of neurology, dermatology, radiation oncology, and ophthalmology (*Figure 2)*. Across our novel template experiments, the generated templates demonstrated a high degree of clinical validity, with 96.2% of required components (475 of 494) deemed clinically appropriate. Moreover, the templates proved highly complete, with our human experts adding on average only 1.2 items per template (24 additions across 20 templates) to achieve full coverage. The templates were not as concise as they could be, with only 73.1% of items (361 of 494) rated by human experts as necessary to include. On average, 18 items per template were rated as necessary, compared to the 8.0 items included in the original templates (for neurology specifically, 23.6 necessary items on average in the generated templates vs 11.6 in the original neurology templates for different conditions).

### 3.2. *Autograder Validation:*

Initial human evaluation of the o3 and GPT-4.1 autograders in a randomized fashion demonstrated a clear superiority for the former model (77.1% vs 71.7% perfect matches, and 1.2% vs 7% complete disagreement). Iterative evaluation of the final o3 autograder with an optimized prompt *(Figure 3)* demonstrated a high degree of concordance across three independent clinical reviewers in rating items as appropriately matched (94.1% full agreement, 4.6% partial agreement, and only 1.3% disagreement), as well as in correctly rating items as unmatched (98.2% accuracy with only 1 false negative among 56 missed elements). Inter-rater reliability analysis revealed substantial agreement among human reviewers by exact agreement and Gwet's AC1, but less substantial agreement by Krippendorff's α (86.9% exact agreement, AC1 = 0.904, α = 0.612). When treating partial matches as acceptable, the autograder achieved 98.7% concordance with human experts; when requiring perfect matches only, concordance ranged from 91-98% across reviewers. Partial disagreements tended to relate to the comprehensiveness and specificity of results (for example, asking for "all imaging results" when a specialist specifically requested "MRI").

### 3.3. *Overall Model Performance*

When evaluating for comprehensiveness across the full generated template (*Figure 4)*, o3 demonstrates clearly superior performance with 92.2% overall comprehensiveness, followed by Claude 4 Sonnet and Gemini 2.5 Pro tied at 82.0%, then Kimi K2 at 78.2%. GPT-4o and Llama-3-70B trail at 76.0% and 71.2% respectively. However, this superior performance came at the cost of conciseness, with o3 producing 2.58x as many template items as were present in the initial template, followed by 2.07x for Claude, 1.56x for Gemini, 1.50x for Kimi, 1.47x for 4o, and 1.27x for Llama.

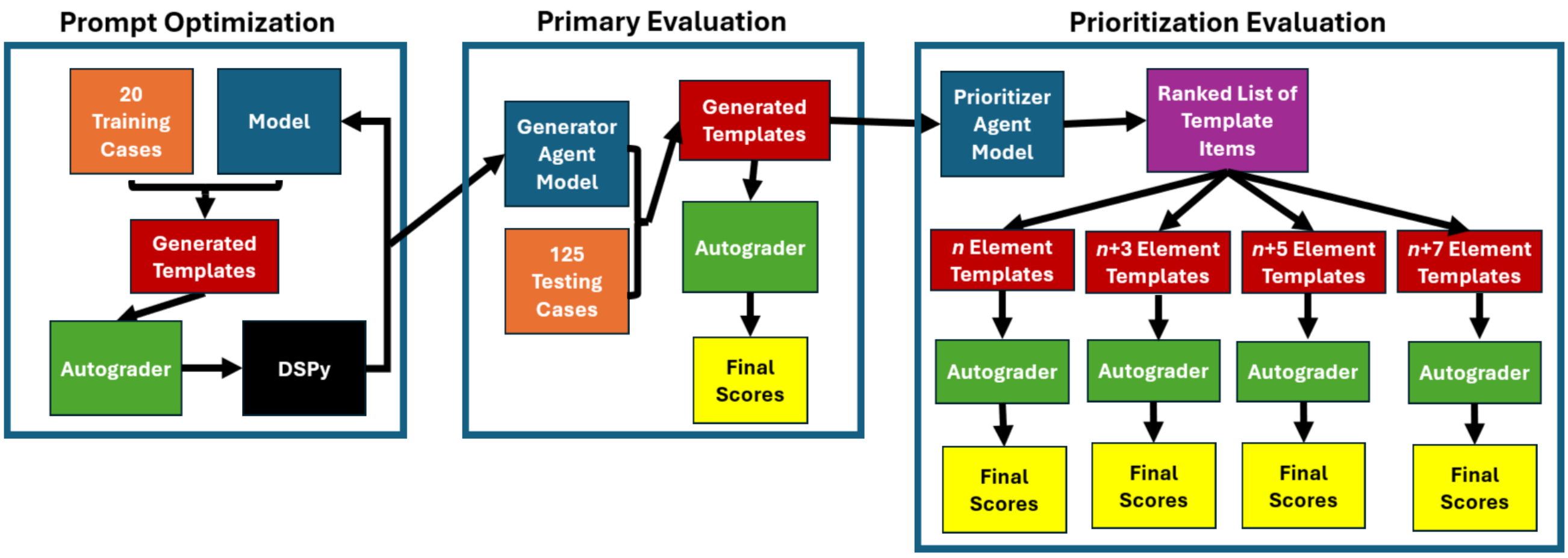

**Figure 1. Overview of Multi-Agent Structure for Template Generation and Evaluation.** Our pipeline comprises three stages: (1) Prompt Optimization, where DSPy iteratively refines a few-shot prompt on a small training set of 20 cases by generating templates, scoring them with an o3 autograder for missing elements or verbosity, and adjusting the prompt until scores converge; (2) Primary Evaluation, in which the finalized prompt is applied to 125 held-out cases and autograded to produce quality and schema-adherence metrics; and (3) Prioritization Evaluation, where a prioritizer agent ranks schema elements and subsets (based on the top-n elements where n equals the number of elements within the original expert template) are assembled and scored to assess the model's ability to surface the most critical items

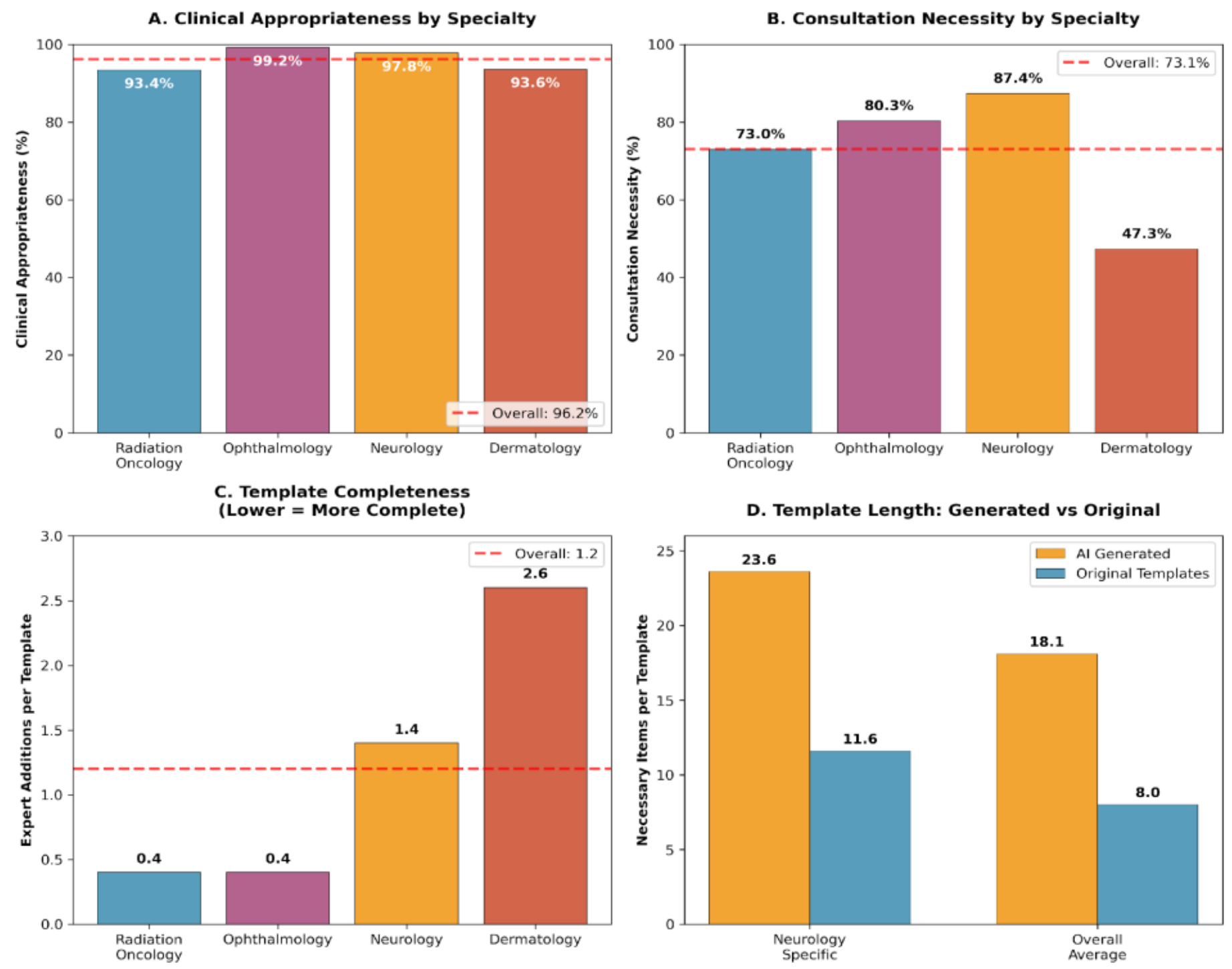

**Figure 2. Expert Clinical Validation of Novel AI-Generated Medical Templates.** All templates generated by o3 with DSPy optimized prompt (A) Clinical appropriateness by specialty showing percentage of template components deemed clinically reasonable by expert reviewers. (B) Consultation necessity by specialty, showing the percentage of components considered essential for specialist consultation. (C) Template completeness is measured by expert additions per template. (D) Template length comparison between AI-generated templates and original Stanford templates, showing necessary items per template for neurology specifically (as there are neuro templates in the original dataset) and the overall average, compared to all specialties.

### 3.4. *Prioritization Performance:*

While OpenAI's reasoning model o3 had the highest score at full length (92.2% coverage), it demonstrated poor prioritization with only 60.4% coverage when confined to the top-*n* items (where n equals the number of items in the original template) (*Figure 5)*. Kimi K2, a recent large-parameter non-reasoning LLM, demonstrated clearly superior performance in this conciseness evaluation, achieving 69.5% coverage at top-n and 77.7% coverage at top-n+3, despite top results of only 78.2%, representing the best retention of performance when forced to prioritize.

### 3.5. *Specialty-Specific Performance:*

Across the full-length evaluation (*Figure 4C*), coverage spanned nearly 35 percentage points: nephrology (97.4%), endocrinology (95.4%), infectious disease (93.9%) and cardiology (93.1%) sat at the top of the heat-map, whereas psychiatry (61.5%), pain medicine (68.2%), sleep medicine (73.1%), and LGBTQ+ medicine (61.4%) clustered at the low end. When the same outputs were truncated to the top-n items (*Figure 5C*), the ordering persisted but the gaps widened: high-performing specialties still retained approximately 70% of reference items, while psychiatry and pain medicine fell below 45%. o3 particularly underperformed relative to its robust performance elsewhere with 14% for LGBTQ+ and 36% for psychiatry. Assessment of the novel templates (*Figure 2B*) demonstrated notably worse conciseness and completeness for the dermatology templates, although the included information was highly appropriate.

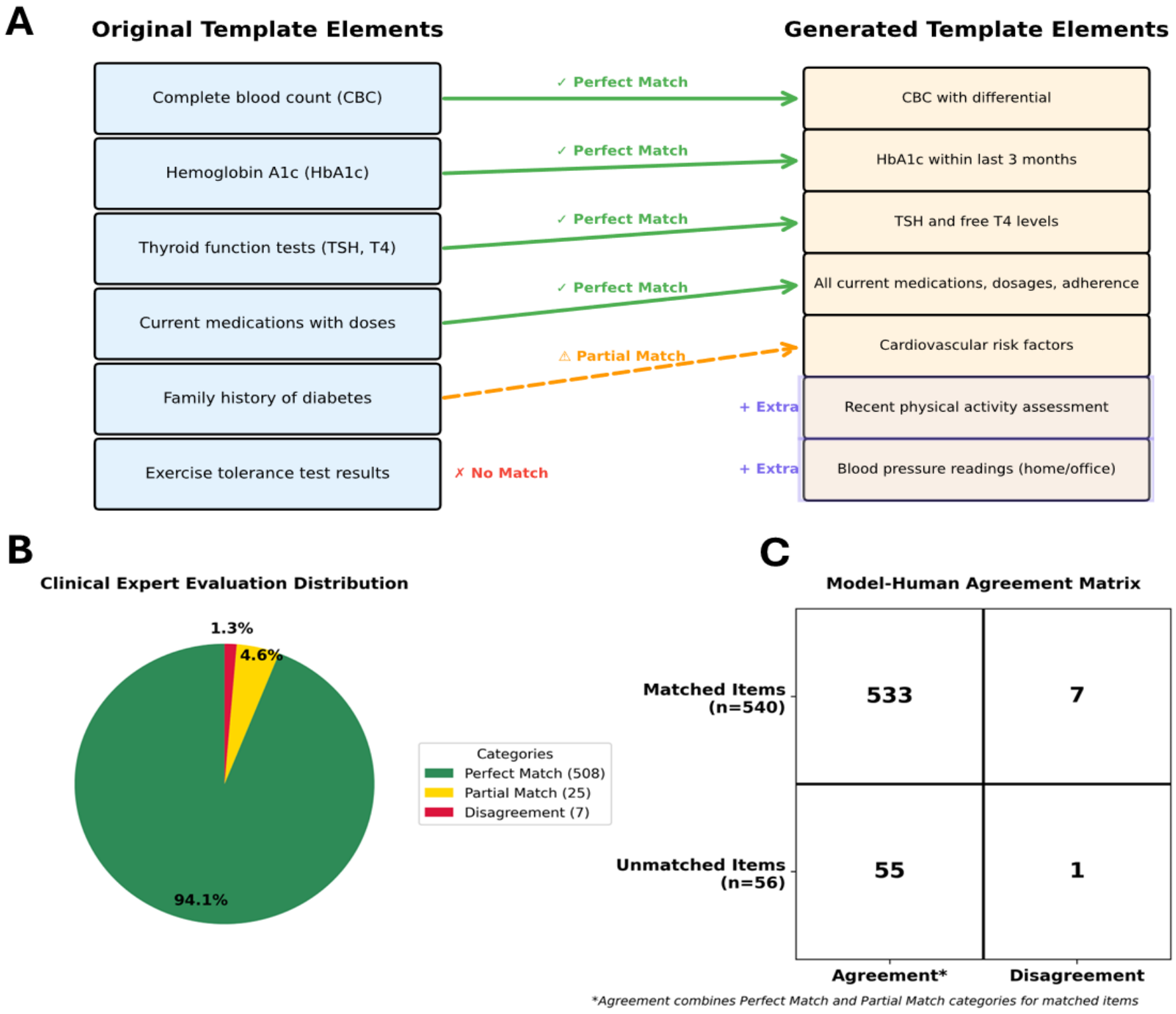

**Figure 3. Structure and Validation of Template Matching Autograder** (A) Example alignment of expert-template elements to a model-generated template, for diabetes mellitus. Arrows represent autograder matches, with color representing human rater assessment of full vs partial match. (B) Distribution of clinician ratings for matched items. (C) Agreement matrix comparing the autograder with human reviewers for both matched and unmatched elements; "agreement" collapses perfect and partial matches

## 4. *Discussion:*

Our study demonstrates that LLMs can generate clinically comprehensive consultation templates that closely approximate expert-created schemas but remain limited in their ability to prioritize high-yield information under real-world constraints. The top-performing model, o3 (a reasoning model from OpenAI), achieved 92.2% coverage of expert template items when unconstrained, but retained only 60.4% coverage when forced to prioritize to the original template length. In contrast, Kimi K2 (a large, open-source non-reasoning model) demonstrated lower full-length comprehensiveness (78.2%) but stronger prioritization retention (69.5%). Across specialties, performance varied substantially: structured domains like nephrology and infectious disease maintained high coverage, while narrative-heavy fields such as psychiatry and pain medicine exhibited significant performance degradation—particularly under length constraints. These findings underscore a central limitation of current LLMs: while they are increasingly capable of enumerating relevant clinical content, they struggle with the task of effective curation.

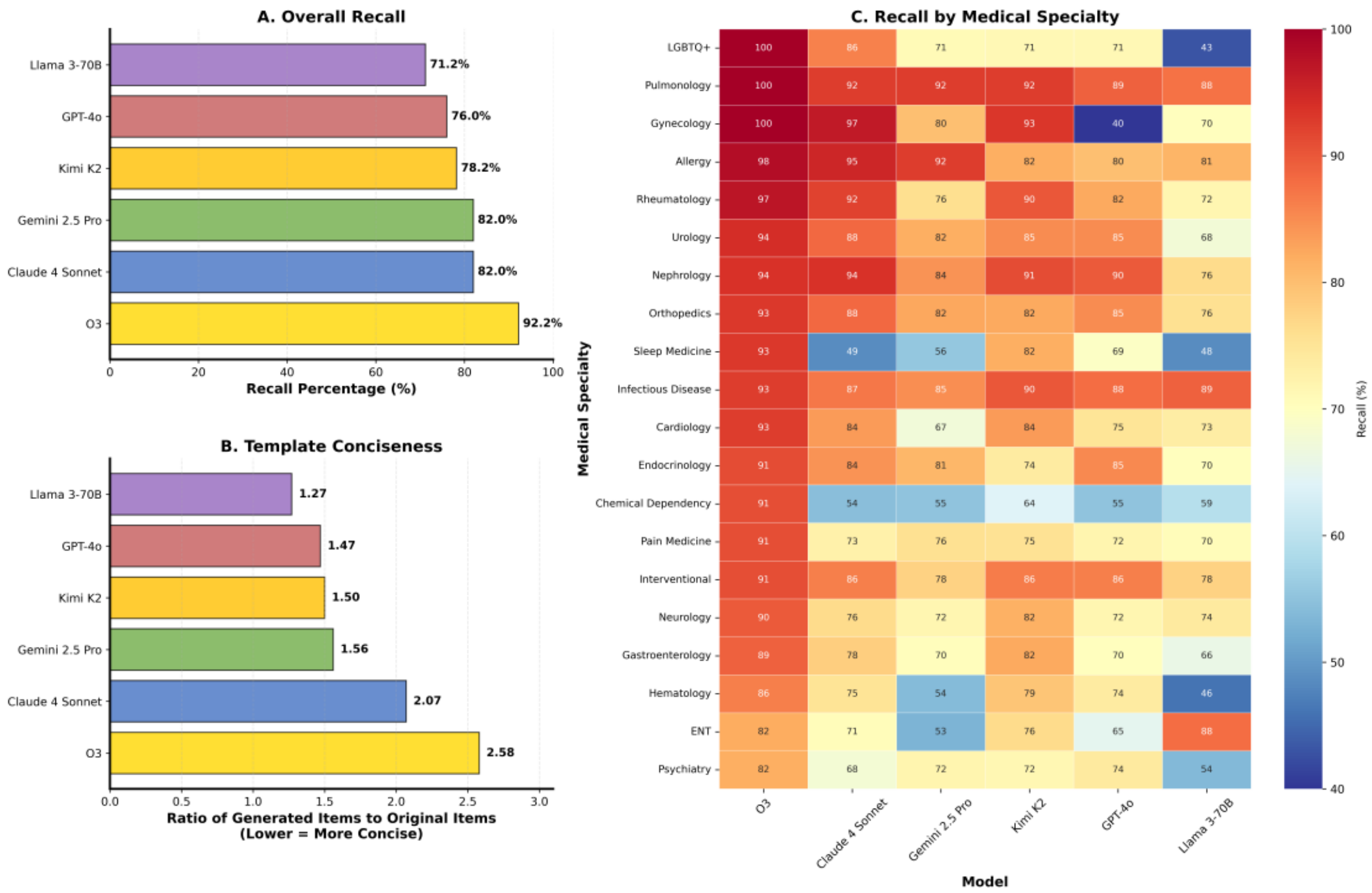

**Figure 4. Multi-Model Performance on Structured Consultation Template Generation.** (A) Overall template item coverage (Recall, in % of reference items matched in generated templates). (B) Conciseness as the excess ratio (generated items ÷ expert template items; lower is more concise). (C) Per-specialty coverage heatmap showing heterogeneity across domains. The central number indicates comprehensiveness scores as a percentage.

While many recent studies have focused on evaluating AI performance within physician-patient interactions and clinical decision making[31–35], the role of LLMs in physician–physician communication remains comparatively underexamined with only limited work on AI-generated handoffs[36] and discharge communications[37,38]. Yet this is precisely where precision, parsimony, and contextual judgment are most acutely required[39]. Unlike patient-facing interactions, which often emphasize rapport, accessibility, or empathic scaffolding[8], physician–physician communication is optimized for signal density and clinical relevance[9]. Where models have been assessed on these tasks, such as in emergency medicine handoffs, similar failures of curation and conciseness have already been observed[36].

The cognitive demands of physician-physician communication are not well-captured by benchmarks oriented around clinical question answering[32], document summarization[40], or patient interaction[41–43]. Moreover, while healthcare AI benchmark studies rely primarily on simulated data and vignettes[41], the templates used in this study reflect real-world specialist information needs and are both actively updated and used in patient care, making them effective for assessing whether LLMs can match expert judgment in structuring clinical communication. Although recent work shows that language models can now identify their own knowledge gaps—either by steering retrieval toward the missing fact[44] or by posing sharper follow-up questions that narrow diagnostic uncertainty[45] —these advances typically occur in closed-loop settings, where the model both detects and benefits from the additional context. A consultation note, by contrast, is written for another clinician, whose attention span, cognitive load, and clinical priorities—not the model's context window—set the real limits on relevance. The ability to recognize what else is needed is, therefore, necessary but not sufficient: models must also curate and compress information in ways that align with the recipient's constraints. For these reasons, our study offers not only a novel task but a clinically grounded benchmark for evaluating language model performance in real-world medical communication.

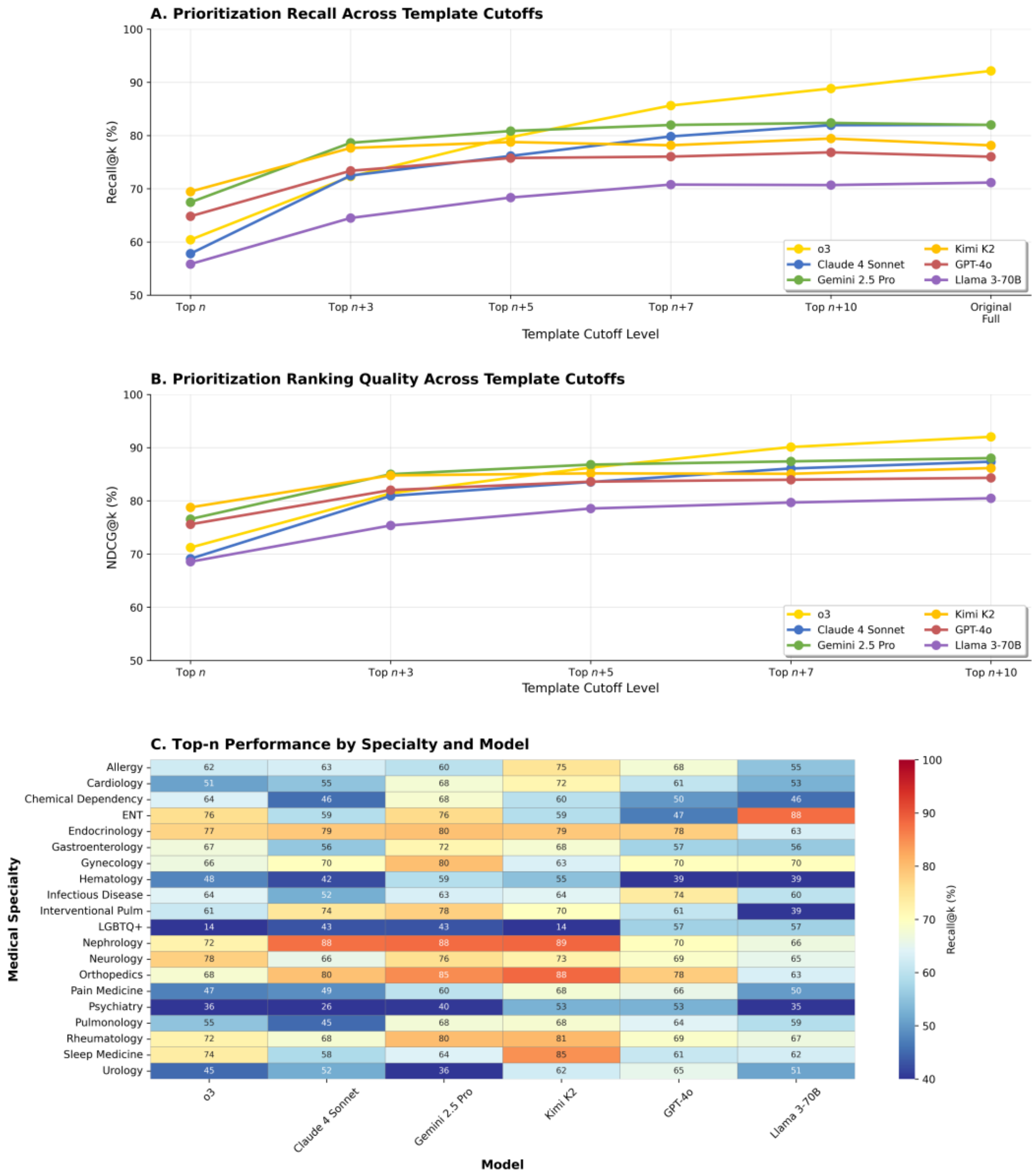

**Figure 5. Prioritization Performance Across Template Cutoffs** (A) Model performance at various cutoffs for template size, based on the prioritizer agent's ranking (using the same underlying model) of template items (n = number of items in the original template, recall@k = the number of original items matched at a given threshold) (B). Analysis of Normalized Discounted Cumulative Gain (NDCG) at a given position in the list, with "required" items valued at 2, and "optional" items valued at 1 (C) Specialty-by-specialty analysis of model performance at top-n cutoff. The central number indicates recall scores.

Even at low prioritization levels, the longest o3-generated templates continued to surface clinically relevant information, with performance improving beyond the top-n+10 items—unlike other models, which plateaued by n+5. These differences reveal deeper epistemological fault lines. Structured, laboratory-anchored specialties like nephrology and infectious disease, which rely on codified data, align well with LLM recall and show minimal degradation under truncation. In contrast, narrative-heavy fields such as psychiatry, pain medicine, and sleep medicine depend on contextual nuance and psychosocial insight. These features are not only harder for models to generate but are also the first to be discarded under compression. That these specialties exhibit both the lowest baseline performance and the steepest prioritization drop-off highlights a core weakness: current LLMs still struggle to assign salience to soft-signal clinical features, even when they can enumerate them in an unconstrained setting.

Expert evaluation of the novel AI-generated templates revealed a consistent pattern: specialists rated a greater number of items as necessary compared to those included in the original human-generated templates. Yet those original templates were themselves authored by specialists, explicitly instructed to balance informational value with the cognitive and workflow constraints of generalist users, and the cognitive load of the receiving specialist. The discrepancy highlights a key issue: in the absence of such constraint, both models and human specialist reviewers tend to favor comprehensiveness over conciseness[7], even though this may be operationally burdensome for generalists in practice[46]. Solutions such as length-aware prompting, structured prioritization, progressive disclosure through expandable sections[47], and integration of routinely available EHR data[15] may thus better align models with real-world clinician workflows. Ultimately, conciseness should not be seen as compromising clinical rigor but rather as essential to avoiding the upstream transfer of documentation burdens to generalist providers.

These tensions are directly analogous to the classical precision–recall trade-offs in information retrieval and detection—tightening decision thresholds typically raises precision at the expense of recall—and are well characterized in methodological work on PR vs. ROC analysis and cost-sensitive decision-making[24,48,49]. While our autograder was calibrated to align closely with expert judgment, it ultimately reflects a single, schema-bound notion of alignment. As generative clinical AI systems grow more capable, the field will require evaluation frameworks that can account for ambiguity, tolerate diversity in clinically reasonable output, and move beyond rigid concordance toward assessments grounded in pragmatic utility.

Beyond generation, ongoing work within the SAGE Consult ecosystem aims to operationalize these templates by auto-populating them with relevant data from the medical record and evaluating how their structure shapes downstream specialist response quality. While automation may reduce the cognitive burden of verbose templates, it cannot fully resolve issues of prioritization or signal-to-noise. A central objective is to assess not just whether templates are clinically coherent, but whether they measurably improve communication efficiency, diagnostic precision, and decision-making accuracy. Parallel efforts focus on empowering generalist clinicians to formulate higher-quality consultation questions—scaffolding the initial request to elicit more targeted and actionable input. We also aim to examine the broader impact of these automations—on workflow, satisfaction, and perceived utility—for both consultants and referring clinicians. Taken together, these initiatives reflect a broader ambition: to move beyond passive documentation toward interactive, intelligence-augmented clinical dialogue, in which both ends of the consultation are meaningfully enhanced by LLM-mediated structure.

This work also has several limitations which are important to acknowledge. These templates are Stanford-specific and may not fully generalize to other healthcare contexts. Further, they are the product of a small number of specialists and may reflect individualized idiosyncrasies in the absence of a universally defined “gold standard”. This task also reflects the inherent limitations of concordance-based assessment: just because a model output does not align with the original template does not necessarily mean it is incorrect. In some cases, a divergent response may be clinically reasonable—or even preferable—depending on context, specialty norms, or emerging practices. Alignment, as measured here, is a pragmatic proxy for quality, not a definitive judgment, and is difficult to reduce to binary metrics in complex clinical circumstances. As such, this benchmark should be viewed as a first-layer filter, useful for surfacing patterns and relative model performance, but insufficient on its own to determine the optimal or most expert-like consultation schema.

Deeper evaluation (through prospective trials and ultimate clinical implementation) will be necessary to fully understand which outputs best serve the needs of generalist-specialist communication. Further technical analyses would strengthen best-practice guidance for template generation. In particular, useful analyses would include (i) ablations with and without DSPy on a model-by-model basis to assess incremental benefit, (ii) a comparison of post-hoc prioritization versus constrained generation at fixed budgets (e.g., n, n+3, n+5), and (iii) detailed prompt tuning to enable a range of comprehensiveness/conciseness equilibria. Techniques for further improving grader robustness, such as multi-LLM ensembling and targeted red-teaming with semantically similar but clinically distinct items, may be beneficial. Further implementation of this work could itself improve salience: by tracking which template elements are emphasized or ignored by both consultants and consultors in real time, systems could learn what information is most valuable. Coupled with EHR auto-population of routine data, this feedback loop may better align model outputs with the practical priorities of daily clinical workflows. Owing to compute and human-review bandwidth limits, we defer these to a follow-on study and to pre-deployment hardening.

In many ways, the consultation template embodies a traditional medical informatics philosophy[50]: structured and deliberately constrained to minimize ambiguity. This approach offers clear advantages in terms of accuracy, interpretability, and interoperability[51]. Yet it also imposes limits, both on expressiveness and on adaptability to novel cases[51]. Large language models, by contrast, offer the promise of flexibility: they can process unstructured input, accommodate a wide range of clinical styles, and synthesize across disparate data. But this flexibility can come at the cost of precision, especially when reasoning implicitly or generating without clear grounding. Effective clinical AI may require combining approaches: leveraging LLMs' broad medical knowledge while imposing external structure through traditional informatics methods[52], grounding tool use[53], or human oversight. Template generation thus illustrates a broader challenge in clinical informatics—determining when to rely on models' flexibility versus when to enforce rigid schemas. The path forward likely involves systems where human expertise defines the structure and constraints, while LLMs help populate and adapt content within those boundaries.

## 5. Conclusion:

This study finds that contemporary large language models can generate comprehensive clinical consultation templates, but struggle with the selective prioritization essential for efficient physician-to-physician communication. Using Stanford's expert-created, actively deployed clinical consultation templates as a benchmark, and employing a physician-calibrated autograder that achieved high concordance with clinical judgment, we reveal that while frontier models successfully capture most clinically relevant information, they produce substantially longer outputs than expert-crafted templates. Moreover, this difficulty with prioritization varies markedly across specialties, with models performing particularly poorly in domains requiring nuanced contextual judgment and communication like psychiatry and pain medicine. Effective integration of LLMs into physician communication will require models that can not only recall relevant information but also recognize what matters most—within the time, attention, and workflow constraints of clinical practice.

| *Original Template* | *O3-Generated Template* |
| --- | --- |
| ### My Clinical Question<br>**My Clinical Question (condition: diabetes mellitus):** ***<br>### My most current assessment<br>**My most current assessment of this problem can be found in the note dated:** {Today/***}<br>### In your clinical question, or current note, please include information on:<br>#### Assessments - Required<br>**Is the patient currently taking medication to treat diabetes?** If yes, please describe.<br>**Is this a new diagnosis of diabetes mellitus?** If not, when was the patient diagnosed with diabetes?<br>**What type of diabetes mellitus does the patient have?** (Type I or Type II)<br>**Insulin (including pump) or non-insulin dependent**<br>#### Assessments - Optional<br>**Diabetic complications** (e.g., CKD, CHF, CAD)<br>**Glucose logs** (if available)<br>#### Diagnostics - Required<br>**Hemoglobin A1c (HbA1c)**<br>**Fasting Lipid Profile**<br>**Fasting Glucose**<br>**Liver Function Tests**<br>**Basic Metabolic Panel**<br>#### Diagnostics - Optional<br>**GAD antibodies**<br>**C-peptide level** | ### My Clinical Question<br>**My Clinical Question (condition: diabetes mellitus):** ***<br>### My most current assessment<br>**My most current assessment of this problem can be found in the note dated:** {Today/***}<br>### In your clinical question, or current note, please include information on:<br>#### Assessments - Required<br>**Type, duration, age at diagnosis**<br>**All current medications, doses, frequency, adherence, recent changes**<br>**Self-monitoring or CGM data:** fasting, post-prandial, patterns<br>**Severe or symptomatic hypoglycemia episodes** (frequency, triggers, treatment)<br>**DKA or hyperosmolar events within past year**<br>**Dietary pattern and carbohydrate counting or exchanges**<br>**Physical activity:** type, minutes/week, barriers; **Weight, BMI, recent change, waist circumference**; **Blood pressure readings, home and clinic**<br>**Microvascular status:** nephropathy, retinopathy, neuropathy findings<br>**Macrovascular disease:** CAD, cerebrovascular, peripheral arterial<br>**Foot exam findings:** pulses, sensation, skin, ulcers<br>**Psychosocial factors:** depression, health literacy, support, financial barriers<br>**Family history of diabetes or cardiovascular disease**<br>#### Assessments - Optional<br>**Pregnancy plans or current pregnancy status**; **Sleep quality, suspected/diagnosed obstructive sleep apnea**; **Alcohol, tobacco, recreational drug use**<br>#### Diagnostics - Required<br>**Hemoglobin A1c** (within last 3 months)<br>**Complete Metabolic Panel (CMP,** within last 6 months)<br>**Urine Albumin-to-Creatinine Ratio** (within last year)<br>**Lipid Panel** (within last year)<br>#### Diagnostics - Optional<br>**C-Peptide or islet antibodies** if type unclear<br>**TSH and thyroid antibodies** (Type 1 or symptoms)<br>**Vitamin B12 level** if on metformin >4 years<br>**Retinal imaging report** (within past year)<br>**12-lead ECG** (if cardiovascular risk factors) |

**Supplementary Figure S1. Example Templates for Endocrinology Diabetes Mellitus Consults**

## 6. References:

1. Kohane IS, Beam AL, Manrai AK. Why Medicine Must Become a Knowledge-Processing Discipline. *NEJM AI*. 2024;1(7):AIp2400582. doi:10.1056/AIp2400582

2. Humphrey KE, Sundberg M, Milliren CE, Graham DA, Landrigan CP. Frequency and Nature of Communication and Handoff Failures in Medical Malpractice Claims. *J Patient Saf*. 2022;18(2):130-137. doi:10.1097/PTS.0000000000000937

3. Starmer AJ, Spector ND, Srivastava R, et al. Changes in Medical Errors after Implementation of a Handoff Program. *N Engl J Med*. 2014;371(19):1803-1812. doi:10.1056/NEJMsa1405556

4. Starmer AJ, Sectish TC, Simon DW, et al. Rates of medical errors and preventable adverse events among hospitalized children following implementation of a resident handoff bundle. *JAMA*. 2013;310(21):2262-2270. doi:10.1001/jama.2013.281961

5. Umberfield E, Ghaferi AA, Krein SL, Manojlovich M. Using Incident Reports to Assess Communication Failures and Patient Outcomes. *Jt Comm J Qual Patient Saf*. 2019;45(6):406-413. doi:10.1016/j.jcjq.2019.02.006

6. Wåhlberg H, Valle PC, Malm S, Broderstad AR. Impact of referral templates on the quality of referrals from primary to secondary care: a cluster randomised trial. *BMC Health Serv Res*. 2015;15:353. doi:10.1186/s12913-015-1017-7

7. Tobin-Schnittger P, O'Doherty J, O'Connor R, O'Regan A. Improving quality of referral letters from primary to secondary care: a literature review and discussion paper. *Prim Health Care Res Dev*. 2018;19(3):211-222. doi:10.1017/S1463423617000755

8. Burden M, Sarcone E, Keniston A, et al. Prospective comparison of curbside versus formal consultations. *J Hosp Med*. 2013;8(1):31-35. doi:10.1002/jhm.1983

9. Lee MS, Ray KN, Mehrotra A, Giboney P, Yee HF Jr, Barnett ML. Primary Care Practitioners' Perceptions of Electronic Consult Systems: A Qualitative Analysis. *JAMA Intern Med*. 2018;178(6):782-789. doi:10.1001/jamainternmed.2018.0738

10. Liddy C, Drosinis P, Fogel A, Keely E. Prevention of delayed referrals through the Champlain BASE eConsult service. *Can Fam Physician*. 2017;63(8):e381-e386.

11. Peeters KMM, Reichel LAM, Muris DMJ, Cals JWL. Family Physician–to–Hospital Specialist Electronic Consultation and Access to Hospital Care: A Systematic Review. *JAMA Netw Open*. 2024;7(1):e2351623. doi:10.1001/jamanetworkopen.2023.51623

12. Arora A, Fekieta R, Spatz E, et al. Implementation and evaluation of an electronic consult program at a large academic health system. *PLOS ONE*. 2024;19(9):e0310122. doi:10.1371/journal.pone.0310122

13. New Frontiers in Team Science: Empowering Patients With AI-Driven E-Consults | DoM Annual Reports. Accessed August 1, 2025. https://domannualreports.stanford.edu/new-frontiers-in-team-science-empowering-patients-with-ai-driven-e-consults/

14. Vimalananda VG, Gupte G, Seraj SM, et al. Electronic consultations (e-consults) to improve access to specialty care: A systematic review and narrative synthesis. *J Telemed Telecare*. 2015;21(6):323-330. doi:10.1177/1357633X15582108

15. Laing S, Jarmain S, Elliott J, et al. Codesigned standardised referral form: simplifying the complexity. *BMJ Health Care Inform*. 2024;31(1):e100926. doi:10.1136/bmjhci-2023-100926

16. OpenAI o3 and o4-mini System Card. Accessed August 1, 2025. https://openai.com/index/o3-o4-mini-system-card/

17. Introducing GPT-4.1 in the API. Accessed August 1, 2025. https://openai.com/index/gpt-4-1/

18. Zapf A, Castell S, Morawietz L, Karch A. Measuring inter-rater reliability for nominal data – which coefficients and confidence intervals are appropriate? *BMC Med Res Methodol*. 2016;16:93. doi:10.1186/s12874-016-0200-9

19. Hayes AF, Krippendorff K. Answering the Call for a Standard Reliability Measure for Coding Data. *Commun Methods Meas*. 2007;1(1):77-89. doi:10.1080/19312450709336664

20. Gwet K. *Handbook of Inter-Rater Reliability: How to Estimate the Level of Agreement Between Two or Multiple Raters*. STATAXIS Publishing Company; 2001.

21. Wongpakaran N, Wongpakaran T, Wedding D, Gwet KL. A comparison of Cohen's Kappa and Gwet's AC1 when calculating inter-rater reliability coefficients: a study conducted with personality disorder samples. *BMC Med Res Methodol*. 2013;13:61. doi:10.1186/1471-2288-13-61

22. Krippendorff K. *Content Analysis: An Introduction to Its Methodology*. Fourth Edition. SAGE Publications, Inc.; 2025. doi:10.4135/9781071878781

23. Khattab O, Singhvi A, Maheshwari P, et al. DSPy: Compiling Declarative Language Model Calls into Self-Improving Pipelines. *arXiv*. Preprint posted online October 5, 2023. doi:10.48550/arXiv.2310.03714

24. Shani G, Gunawardana A. Evaluating Recommendation Systems. In: Ricci F, Rokach L, Shapira B, Kantor PB, eds. *Recommender Systems Handbook*. Springer US; 2011:257-297. doi:10.1007/978-0-387-85820-3_8

25. Järvelin K, Kekäläinen J. Cumulated gain-based evaluation of IR techniques. *ACM Trans Inf Syst*. 2002;20(4):422-446. doi:10.1145/582415.582418

26. OpenAI, Hurst A, Lerer A, et al. GPT-4o System Card. *arXiv*. Preprint posted online October 25, 2024. doi:10.48550/arXiv.2410.21276

27. Gemini 2.5: Our most intelligent AI model. Google. March 25, 2025. Accessed August 1, 2025. https://blog.google/technology/google-deepmind/gemini-model-thinking-updates-march-2025/

28. Introducing Claude 4. Accessed August 1, 2025. https://www.anthropic.com/news/claude-4

29. Kimi K2: Open Agentic Intelligence. Accessed August 1, 2025. https://moonshotai.github.io/Kimi-K2/

30. Introducing Meta Llama 3: The most capable openly available LLM to date. Accessed August 1, 2025. https://ai.meta.com/blog/meta-llama-3/

31. Brodeur PG, Buckley TA, Kanjee Z, et al. Superhuman performance of a large language model on the reasoning tasks of a physician. *arXiv*. Preprint posted online June 2, 2025. doi:10.48550/arXiv.2412.10849

32. Singhal K, Tu T, Gottweis J, et al. Toward expert-level medical question answering with large language models. *Nat Med*. 2025;31(3):943-950. doi:10.1038/s41591-024-03423-7

33. Tu T, Palepu A, Schaekermann M, et al. Towards Conversational Diagnostic AI. *arXiv*. Preprint posted online January 11, 2024. doi:10.48550/arXiv.2401.05654

34. Goh E, Gallo R, Hom J, et al. Large Language Model Influence on Diagnostic Reasoning. *JAMA Netw Open*. 2024;7(10):e2440969. doi:10.1001/jamanetworkopen.2024.40969

35. Busch F, Hoffmann L, Rueger C, et al. Current applications and challenges in large language models for patient care: a systematic review. *Commun Med*. 2025;5:26. doi:10.1038/s43856-024-00717-2

36. Hartman V, Zhang X, Poddar R, et al. Developing and Evaluating Large Language Model–Generated Emergency Medicine Handoff Notes. *JAMA Netw Open*. 2024;7(12):e2448723. doi:10.1001/jamanetworkopen.2024.48723

37. Williams CYK, Subramanian CR, Ali SS, et al. Physician- and Large Language Model-Generated Hospital Discharge Summaries. *JAMA Intern Med*. 2025;185(7):818-825. doi:10.1001/jamainternmed.2025.0821

38. Landman AB, Tilak SS, Walker GA. Artificial Intelligence–Generated Emergency Department Summaries and Hospital Handoffs. *JAMA Netw Open*. 2024;7(12):e2448729. doi:10.1001/jamanetworkopen.2024.48729

39. McCoy LG, Manrai AK, Rodman A. Large Language Models and the Degradation of the Medical Record. *N Engl J Med*. Published online October 31, 2024. doi:10.1056/NEJMp2405999

40. Ntinopoulos V, Rodriguez Cetina Biefer H, Tudorache I, et al. Large language models for data extraction from unstructured and semi-structured electronic health records: a multiple model performance evaluation. *BMJ Health Care Inform*. 2025;32(1):e101139. doi:10.1136/bmjhci-2024-101139

41. Bedi S, Liu Y, Orr-Ewing L, et al. Testing and Evaluation of Health Care Applications of Large Language Models. *JAMA*. 2025;333(4):319-328. doi:10.1001/jama.2024.21700

42. Hager P, Jungmann F, Holland R, et al. Evaluation and mitigation of the limitations of large language models in clinical decision-making. *Nat Med*. 2024;30(9):2613-2622. doi:10.1038/s41591-024-03097-1

43. McCoy LG, Swamy R, Sagar N, et al. Do Language Models Think Like Doctors? *medRxiv*. Preprint posted online February 12, 2025:2025.02.11.25321822. doi:10.1101/2025.02.11.25321822

44. Wang K, Duan F, Li P, Wang S, Cai X. LLMs Know What They Need: Leveraging a Missing Information Guided Framework to Empower Retrieval-Augmented Generation. *arXiv*. Preprint posted online April 22, 2024. doi:10.48550/arXiv.2404.14043

45. Li SS, Mun J, Brahman F, Ilgen JS, Tsvetkov Y, Sap M. Aligning LLMs to Ask Good Questions A Case Study in Clinical Reasoning. *arXiv*. Preprint posted online February 20, 2025. doi:10.48550/arXiv.2502.14860

46. Baines R, Tredinnick-Rowe J, Jones R, Chatterjee A. Barriers and Enablers in Implementing Electronic Consultations in Primary Care: Scoping Review. *J Med Internet Res*. 2020;22(11):e19375. doi:10.2196/19375

47. Panigutti C, Beretta A, Fadda D, et al. Co-design of Human-centered, Explainable AI for Clinical Decision Support. *ACM Trans Interact Intell Syst*. 2023;13(4):21:1-21:35. doi:10.1145/3587271

48. Davis J, Goadrich M. The relationship between Precision-Recall and ROC curves. In: *Proceedings of the 23rd International Conference on Machine Learning*. ICML '06. Association for Computing Machinery; 2006:233-240. doi:10.1145/1143844.1143874

49. Elkan C. The foundations of cost-sensitive learning. In: *Proceedings of the 17th International Joint Conference on Artificial Intelligence - Volume 2*. IJCAI'01. Morgan Kaufmann Publishers Inc.; 2001:973-978.

50. Bates DW, Kuperman GJ, Wang S, et al. Ten commandments for effective clinical decision support: making the practice of evidence-based medicine a reality. *J Am Med Inform Assoc JAMIA*. 2003;10(6):523-530. doi:10.1197/jamia.M1370

51. Rosenbloom ST, Denny JC, Xu H, Lorenzi N, Stead WW, Johnson KB. Data from clinical notes: a perspective on the tension between structure and flexible documentation. *J Am Med Inform Assoc JAMIA*. 2011;18(2):181-186. doi:10.1136/jamia.2010.007237

52. Geng S, Cooper H, Moskal M, et al. JSONSchemaBench: A Rigorous Benchmark of Structured Outputs for Language Models. *arXiv*. Preprint posted online February 27, 2025. doi:10.48550/arXiv.2501.10868

53. Goodell AJ, Chu SN, Rouholiman D, Chu LF. Large language model agents can use tools to perform clinical calculations. *Npj Digit Med*. 2025;8(1):163. doi:10.1038/s41746-025-01475-8